\documentclass[10pt,twocolumn,letterpaper]{article}

\usepackage{cvpr}              

\usepackage{graphicx}
\usepackage{amsmath}
\usepackage{amssymb}
\usepackage{booktabs}
\usepackage{multirow}

\usepackage[switch,pagewise]{lineno} 
\usepackage[ruled]{algorithm2e}
\usepackage{algorithmic}

\usepackage[pagebackref,breaklinks,colorlinks]{hyperref}

\usepackage[capitalize]{cleveref}
\crefname{section}{Sec.}{Secs.}
\Crefname{section}{Section}{Sections}
\Crefname{table}{Table}{Tables}
\crefname{table}{Tab.}{Tabs.}

\def\cvprPaperID{*****} 
\def\confName{CVPR}
\def\confYear{2022}

\begin{document}

\title{3DRM: Pair-wise Relation Module for 3D Object Detection}

\author{Yuqing Lan, Yao Duan, Yifei Shi\\
National University of Defense Technology\\
{\tt\small lanyuqingkd,duanyao16@nudt.edu.cn,yifei.j.shi@gmail.com}\\

\and
Hui Huang\\
Shenzhen University\\
{\tt\small hhzhiyan@gmail.com}
\and
Kai Xu\\
National University of Defense Technology\\
{\tt\small kevin.kai.xu@gmail.com}

}
\maketitle

\begin{abstract}
  Context has proven to be one of the most important factors in object layout reasoning for 3D scene understanding. Existing deep contextual models either learn holistic features for context encoding or rely on pre-defined scene templates for context modeling. We argue that scene understanding benefits from object relation reasoning, which is capable of mitigating the ambiguity of 3D object detections and thus helps locate and classify the 3D objects more accurately and robustly.
  To achieve this, we propose a novel 3D relation module (3DRM) which reasons about object relations at pair-wise levels. The 3DRM predicts the semantic and spatial relationships between objects and extracts the object-wise relation features. We demonstrate the effects of 3DRM by plugging it into proposal-based and voting-based 3D object detection pipelines, respectively. Extensive evaluations show the effectiveness and generalization of 3DRM on 3D object detection. Our source code is available at \url{https://github.com/lanlan96/3DRM}.
\end{abstract}

\label{intro}
\section{Introduction}

3D scene understanding involves the detection of 3D objects and the inference of their spatial layouts. It is one of the most fundamental problems in graphics, vision and robotics. Recently, the fast development of 3D data acquisition and reconstruction techniques has made the collection of large-scale 3D real-world scene data more accessible than ever.
Nowadays, the reconstructed real-world 3D scene datasets (e.g. S3DIS \cite{Armeni20163D} and ScanNet \cite{dai2017}) usually contain a lot of various objects distributing in multiple areas or rooms. This makes 3D object detection quite challenging.

\begin{figure}[!t]
  \centering
  \includegraphics[width=1.0\linewidth]{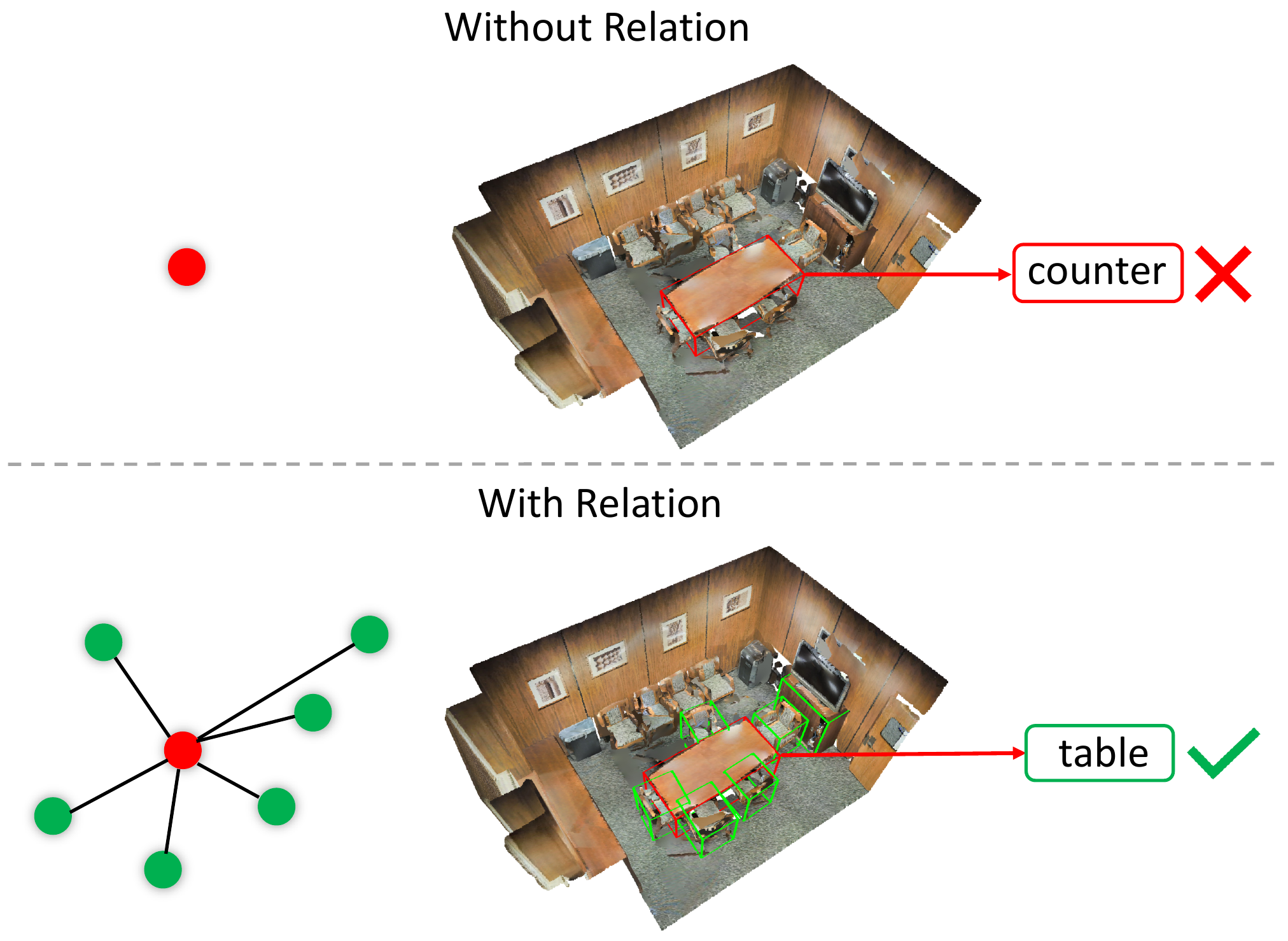}
  \caption{We propose 3DRM which reasons about object relations in 3D object detection. For example, a single object can usually not be correctly identified without knowing the context around it. The proposed 3DRM is able to boost 3D object detection by reasoning the relations between the surrounding objects.}
  \label{fig:teaser}
  \end{figure}

Context in 3D scenes refers to the spatial or semantic relations between different objects, which is critical to scene understanding (see Figure~\ref{fig:teaser}). It has proven to be extremely useful in 3D object detection~\cite{qi20173d,zhang2017deepcontext,shi2016data}.
In the era of deep learning, contextual modeling continues to play an important role in scene analysis~\cite{shi2019hierarchy,Ye_2018_ECCV,Engelmann_2017_ICCV,Porway2008A}. Existing context-based deep learning approaches either extract a holistic feature encoding contextual information via 3D or graph convolutional networks~\cite{li2018pointcnn,qi2017pointnet++,feng2020relation} or learn contextual information from pre-defined templates of object layout \cite{deepcontext}. However, these approaches require large amount of training data with complete scene geometry or object layout, which limits their flexibility.

Qi el al.~\cite{Charles2017PointNet} propose PointNet for learning 3D representations directly from point cloud data to perform classification and segmentation, yielding many follow-up works. These 3D geometric features have so far been the mainstream in scene understanding. However, a powerful method calls for more diverse features, inspired by the success of multi-modal object detection in 2D images~\cite{feng2020deep}. In fact, objects in a particular scene are functionally related or have correlation in structure. Such inherent relations can supply a new type of high-level 3D features which may fill in the gap in 3D object detection. 

\begin{figure*}[htp]

  \centering
    \includegraphics[width=1.0\linewidth]{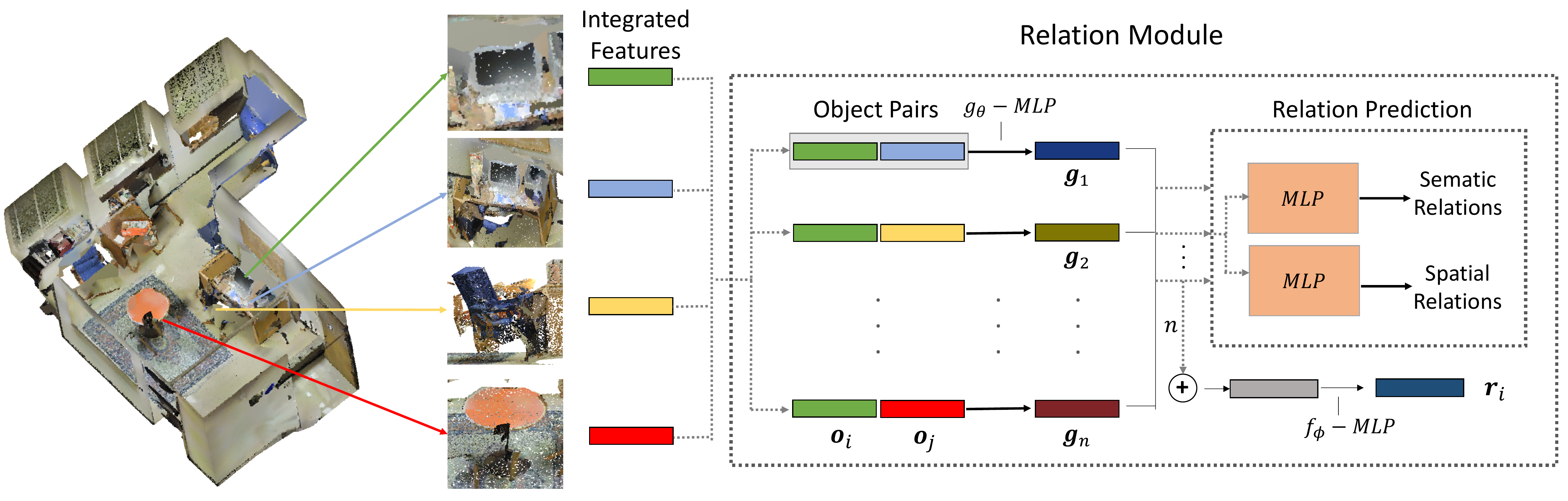}
  \caption{The network architecture of 3DRM. Input of Relation Module comes from features of object candidates extracted by different backbones. Integrated features of different objects are matched by pairs and go through MLP called $g_{\theta}$. Pair-wise Features corresponding to the same object like the green one will be added together and go through another MLP called $f_\phi$ to obtain relation feature $r_i$. Features extracted by $g_{\theta}$ are fed into different MLPs to reason different relations of object pairs. In summary, the Relation Module outputs predictions of semantic and spatial relations as well as relation features $r_i$.}
  \label{fig:fig_relation}       
  \end{figure*}

In this paper, we propose to model object context through reasoning about their relations.
The proposed 3D Relation Module, or \emph{3DRM} for short, operates directly on features of 3D point cloud and outputs the relational features which can be used to boost the performance of various object detection frameworks for 3D scenes (see Figure~\ref{fig:fig_relation}).
The core of our method is a pair-wise relation reasoning module which is not only capable of predicting relational attributes of object pairs, but also mitigates the ambiguity of 3D objects that are hard to detect.
Different from previous works, our method does not rely on pre-defined scene templates for contextual features extraction.

3DRM adapts the Relation Network~\cite{santoro2017simple} to reasoning about relations between object pairs in 3D representations. Objects in indoor scenes are typically semantically and spatially related. Given integrated features extracted by different backbones with scene point cloud as input, 3DRM performs pair-wise object relation reasoning with a relation module. Objects in the same scene are paired using specific matching strategies. Pair-wise object features are then processed by the proposed 3DRM and prediction of relations is performed with extracted relation features which will be leveraged to help the task of detection. 

3DRM is a plug-and-play module which can be applied to different 3D detection frameworks to detect 3D objects more accurately and robustly. We apply 3DRM to two 3D object detection backbones, and evaluate its performance on three challenging datasets. Extensive experiments demonstrate the effectiveness of 3DRM. Specifically, applying 3DRM to different detection backbones achieves $\textbf{30}\%$ improvement on S3DIS \cite{Armeni20163D}, $\textbf{3.8}\%$ on ScanNetV2 \cite{dai2017} and $\textbf{1.4}\%$ on SUN RGB-D dataset~\cite{song2015}.

In summary, we make the following contributions:
\begin{itemize}
\item We propose a 3D relation module which reasons about the relations between 3D objects. Different from other methods which only extract geometry or location features for individual objects, our method is able to capture relation features. This diversifies the feature palette of 3D point cloud and can be combined with other features to boost the performance of object detection.
\item We design four dedicated relationships of semantic and spatial properties between objects which can be computed in real time instead of manual annotation. 

\item Extensive experiments demonstrate the benefits of relation information. We plug our relation module into two popular detection backbones. The results show substantial improvements on the S3DIS, ScanNetV2 and SUN RGB-D datasets which demonstrates that our design is effective and can be widely applicable.
\end{itemize}

\label{relatedwork}
\section{Related Work}

\textbf{3D object detection.} 3D object detection in point cloud is now common in indoor scene understanding~\cite{Lin_2013_ICCV, Charles2017PointNet,qi2017pointnet++,li2018pointcnn,wang2018sgpn,shi2019hierarchy, Xie_2020_CVPR, Chen_2020_CVPR, Qi_2019_ICCV, Hou_2019_CVPR, H3DNet, feng2020relation, Huang_2018_ECCV, zhao2018triangle} and autonomous driving~\cite{Qi_2018_CVPR,Chen2017Multi,ku2018joint,Yang2018PIXOR,Liang2018Deep,shi2019pointrcnn}.

Yang \emph{et al.} \cite{yang2019learning} directly predict object bounding  boxes from a learned global feature vector and obtain instance masks by segmentation points inside a bounding box. VoteNet~\cite{qi2019deep} highlights the challenge of directly predicting bounding box centers in sparse 3D data as most surface points are far away from object centers. Shi \emph{et al.} \cite{shi2019hierarchy} also generate the objects proposals by graph cuts for an over-segmentation of the point cloud based on point normal differences to create the initial set of segments and leverages the features from PointCNN~\cite{li2018pointcnn} to explore the hierarchy structure of objects and context. 3D-MPA~\cite{Engelmann20CVPR} adapts the object-center approach, extends it with a branch for instance mask prediction and replaces NMS with a grouping mechanism of jointly-learned proposal features. However, all these methods take PointNet, PointNet++ or PointCNN as their backbone to extract geometry features which is insufficient. Relationships between objects provide abundant information for scene understanding which is usually ignored. Huang \emph{et al.} \cite{huang2018cooperative} also emphasize the importance of context relations among objects for 3D box estimation.

\textbf{Relation reasoning in 3D.} Since the Relation Network~\cite{santoro2017simple} has been proposed, there has been an explosion of methods that apply the Relation Network \cite{santoro2017simple} in various tasks on 2D image, such as object detection~\cite{Xu_2019_CVPR,hu2018relation,mou2019relation,fan2020fsod,zheng2020foreground}, semantic segmentation~\cite{li2020spatial}, object recognition~\cite{chen2017spatial,zhang2020relation}, action recognition~\cite{Simonyan2014Two,cui2020learning,huang2020improving}, object relationship detection~\cite{Krishna2016Visual,mi2020hierarchical,liu2020beyond}, VQA~\cite{santoro2017simple,Agrawal2017VQA,cadene2019murel,le2020hierarchical}, few-shot learning~\cite{sung2018learning}, scene graph generation~\cite{wang2019exploring} etc.~\cite{hu2018relation} uses a relation module to reason object relations and improves the recognition accuracy.~\cite{mou2019relation} applies relation modules on features extracted from VGG-16 for semantic segmentation in Aerial Scenes. All these work demonstrate the importance of relation reasoning in visual tasks. 

As the result of the great success of relation reasoning in the 2D domain, some work has already explored the relationships in 3D data. \cite{duan2019structural} equips the PointNet++ \cite{qi2017pointnet++} with relation network to reason about the structural dependencies of local regions in 3D point clouds and get a big boost on the tasks of 3D point cloud classification and part segmentation. Liu \emph{et al.} \cite{liu2019relation} propose a convolution operator which encodes geometric relations of points by reasoning about the spatial layout of points for point cloud analysis. \cite{yang2019learning} improves the performance on both 3D object recognition and retrieval tasks, which reinforces the information for individual view by modeling the relationships between its inside regions and the corresponding regions in other views, and then integrates the information from multiple views by modeling the inter-relationships together. \cite{kulkarni20193d} reasons about the relative pose between each pair of objects to improve 3D pose prediction. \cite{zheng2014recurring,huang2015support} use relation graphs or specific relations like \emph{support} to perform relation reasoning towards different components of an object. \cite{song2017web3d} leverages case-based reasoning to measure similarity between different furniture layouts. \cite{huang2016structure} designs five types of relations, which however are dependent on handcrafted labeling as well as time-consuming for relations like \emph{facing}, to build structure graphs of furniture in different scenes respectively, and performs scene matching for novel scene synthesis.

However, there are few work reasoning about the relationships between 3D objects pairs in the indoor scenes by automatic computation and taking advantage of the relationships to capture the relation feature for improving the 3D object detection performance.

\label{methods}
\section{Method}

\begin{figure*}[h]

\centering
  \includegraphics[width=1.0\linewidth]{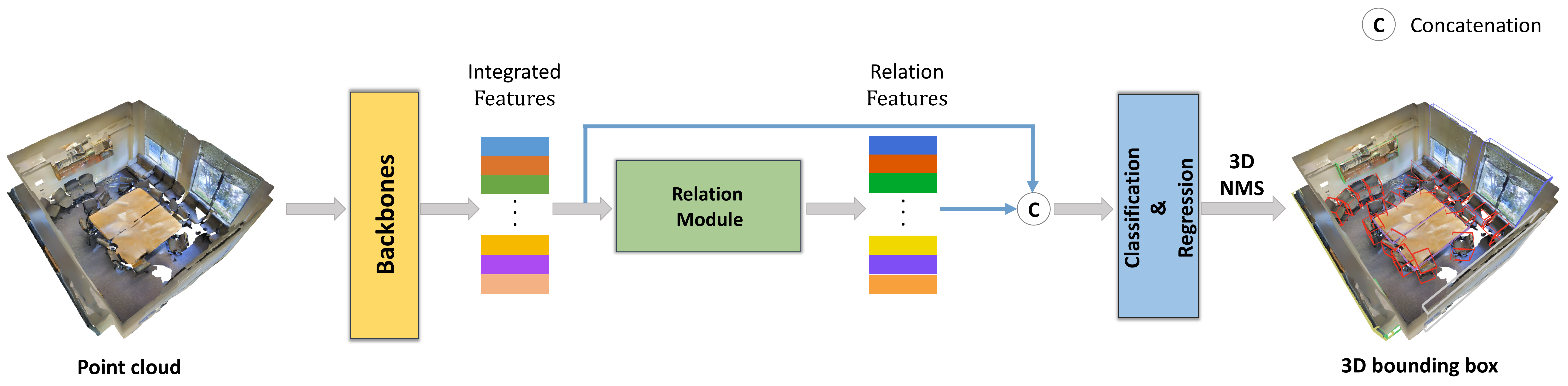}
\caption{The 3D detection pipeline utilizing our 3DRM. Input point clouds go through the backbone feature extraction networks to gain integrated features which are then sent into Relation Module to get the relation features. Both of integrated features and relation features are used to perform classification and regression. 3D non-maximum suppression (NMS) is followed to output the final 3D bounding box.}
\label{fig:framework}       
\end{figure*}

\label{Overview}
\subsection{Overview}
Traditional networks for 3D object detection mainly leverage geometric features of objects to regress the bounding box and conduct classification. Different from that, our Relation Module is aimed for learning the pair-wise objects relationships to extract relation features, which fills a gap in features of 3D data representations.
The goal of this paper is to apply the proposed 3DRM to the existing popular detection pipelines with point clouds as input
and improve the final performance. 

In our method, we leverage two detection frameworks to generate object candidates and extract their features: proposal-based methods and voting-based methods. With features of objects extracted by the backbones as input, our 3DRM can build up pair-wise object relations and extract the comprehensive relation features. As a result, the relation feature will be concatenated with input feature to help the task of detection (Section~\ref{Relation reasoning}). Strategies about application of our Relation Module on different backbones are demonstrated in Section~\ref{Strategies for backbones}. Designs for loss function about different backbones are illustrated in Section~\ref{Loss Function}.

\subsection{Relation module}
\label{Relation reasoning}
Different from existing work on 3D object detection which extract contextual information by taking the entire scene as input, our method learns the object-level relational context features and infers attributes between object pairs. We argue that the relation between object pairs is beneficial to object reasoning for 3D scene understanding. Motivated by the Relation Networks proposed in \cite{hu2018relation}, we adapt the relation module to 3D object detection task.
Unlike the strategy that applies relation prediction on the cells of feature map,
we perform the relation prediction on individual objects, so object relations are explicitly obtained. On the other hand, our goal of relation reasoning is not to aggregate the global context feature for predicting the attributes of the entire scene. Since we aim to detect individual objects, our relation reasoning module is essentially learning the relation-related feature for individual objects.

The architecture of 3DRM is shown in Figure \ref{fig:fig_relation}. The input of our relation module is the object candidates with their features $\textbf{o}_{i} \in \mathbb{R}^{d}$ generated by backbone methods. We propose to use a deep network to extract relational features and predict the relations of a pair of objects.
Specifically, for each object $\textbf{o}_{i}$, we randomly choose $k$ objects $\textbf{o}_{j}$ in the same scene, which then compose several pairs. In Eq. (\ref{equ:rn_feature}), the pairwise function $g_{\theta}$ aims to exploit the semantic or spatial relations between $\textbf{o}_{i}$ and $\textbf{o}_{j}$, and then $f_\phi$ fuses the relations followed by an element-wise sum for all $\textbf{o}_{j}$. As a result, $ \textbf{r}_{i}$ is the learned relational feature of $\textbf{o}_{i}$.

\begin{equation}
\label{equ:rn_feature}
\textbf{r}_{i}  = f_\phi(\sum_{\forall_{j}}g_{\theta}(\textbf{o}_{i},\textbf{o}_{j})), j\in\lbrace1,\ldots,k\rbrace
\end{equation}  
where both $ i $ and $ j $ are the indexes of the objects in the same scene; 
$ g_\theta $ and $ f_\phi $ are the functions implemented by MLPs. 

At the same time, we also predict the relationships of each pairs of objects. The output of the pairwise function $g_{\theta}$ will be sent to the classification MLPs to predict the relation label $l_{{rn}}$. Note that $ h_\varphi $ is the function implemented by MLPs.
  
\begin{equation}
\label{equ:rn}
l_{{rn}}  = h_\varphi(g_{\theta}(\textbf{o}_{i},\textbf{o}_{j}))  ,j\in\lbrace1,\ldots,k\rbrace
\end{equation}

We design different classifiers for different relations. There are four types of relations which include semantic and spatial information: \emph{group}, \emph{same as}, \emph{support} and \emph{hang on}. Relations of different components within an object are omitted since we argue that pair-wise object relations are more significant to indoor object detection. The reason why we choose these four types of relations is that these relations are typical in indoor scenes and sufficient for attaining useful relation features for detection intuitively. For the sake of efficiency and practicability, relations like \emph{facing} or \emph{contain} are not under consideration in this paper because indoor 3D object detection is usually aimed for representative objects like chairs and tables instead of bottles of wine in the cabinet. In what follows, we describe how to compute the relation labels of each pair.

\begin{algorithm}[t]
\caption{Pseudo code for spatial relation formulation.}
  \LinesNumbered
  \label{algorithm1}
  \For{\textbf{all} object pairs $(p_{i},p_{j})$}{
      compute axis-distance $\Psi_x, \Psi_y, \Psi_z$.\\
      compute plane-wise IoU $\Omega_{xy}, \Omega_{xz}, \Omega_{yz}$.\\
      $label_{{rn}} \leftarrow 0$\\
      \If{$\Psi_z \leq \tau_z $ and $\Omega_{xy} > \tau_{xy}$}{
        relation $\leftarrow$ \emph{support}, $label_{{rn}} \leftarrow 1$
      }
      \ElseIf{$\Psi_y \leq \tau_y,\Omega_{xz} > \tau_{xz}$ or $\Psi_x \leq \tau_x,\Omega_{yz} > \tau_{yz}$} {
        relation $\leftarrow$ \emph{hang on}, $label_{{rn}} \leftarrow 1$
      }
  }
\end{algorithm}

\textbf{Semantic relations.} Generally, there are many objects belonging to the same category in the same scene. For example, in most cases, couples of chairs often simultaneously exist in a conference room. We argue that semantic information is beneficial to detection tasks. In this paper, semantic information 
covers two relations: \emph{group} and \emph{same as}.

\emph{Group} relations learn the potential connection between objects that have the same categorical class label and learn the diversity of the different types of objects. We try to capture the relations between objects in terms of the semantic class-specific properties.
Distinguishing semantic class relations from various objects benefits the classification, which is equal to answer the question that whether two objects have the same categorical label or not.

\emph{Same as} relations indicate that a pair of objects 
may belong to the same instance even if they cover different parts of the object. This type of relation gets the candidate objects belonging to same instance closer while keeps other objects away. 
Actually this is an exploration of the instance’s intrinsic property.

\textbf{Spatial relations.} Objects in the same scene are potentially connected in the field of space, especially for those with 3D representations. Spatial information indeed implies abundant and useful information, which is helpful for better understanding of the scene with our relation network. We divide spatial relations into two canonical relations: \emph{support} and \emph{hang on}. The algorithm for spatial relation computation is illustrated in Algorithm~\ref{algorithm1}.

\emph{Support} relations describe the spatially adjacent relations between the objects. In other words, it emphasizes  
that objects are functionally close and connected with this relation. Automatically extracting relations of \emph{support} is quite challenging due to the noisy and partial occlusion of real 3D scans. We define that two objects have relations of \emph{support} only when they are close enough on the z-axis and IoU between 
their projection in the horizontal plane is large enough. Furthermore, if object A is on top of object B and ground projection of object A against the one of object B is higher than a certain threshold, we argue that object A and object B have \emph{support} relations. For example, a flower vase standing on the table describes the relation of \emph{support}.  

Specifically, when deciding the \emph{support} relations of two proposals, we first  check if 1) their 
relative height $\Psi_z(p_{i},p_{j})$ between the lower surface of one object and the upper surface of the other object is smaller than the threshold $\tau_z$, and 2)
the overlapping ratio in xy-plane $\Omega_{xy}(p_i,p_j)$ surpasses the threshold $\tau_{xy}$. If so, they will have relations of \emph{support}.

\begin{equation}
  \label{equ:z-dis}
  \psi_z(p_{i},p_{j})  = \mid \nu_{max}^z(p_i)- \nu_{min}^z(p_j)  \mid 
  \end{equation}

\begin{equation}
  \label{equ:z-dis2}
  \Psi_z(p_{i},p_{j})  = \min(\psi_z(p_{i},p_{j}),\psi_z(p_{j},p_{i})) 
  \end{equation}
where $ p_{i} $ and $ p_{j} $ are two objects, $\nu^z(p_i)$ and $\nu^z(p_j)$ denote their position on z-axis for points of them.  $\Psi_z(p_{i},p_{j}) $ is their minimum distance on the z-axis.

\begin{equation}
  \label{equ:z-dis3}
  \Omega_{xy}(p_i,p_j)  = \max(\frac{\delta_{xy}(p_i,p_j)}{\beta_{xy}(p_i)}, \frac{\delta_{xy}(p_i,p_j)}{\beta_{xy}(p_j)}) 
  \end{equation}
where $\delta_{xy}(\cdot)$ computes the intersection area of projection for two objects. $\beta_{xy}(\cdot)$ denotes the size of projection area. $\Omega_{xy}(p_i,p_j)$ indicates the larger overlapping ratio of the IoU towards these two projection areas.

\emph{Hang on} relations imply that two proposals are horizontally adjacent and one hangs on the other. Similar to relations of \emph{support}, if object A is horizontally close to object B and perpendicular projection (parallel to xz-plane or yz-plane) of object A against the object B is higher than a certain threshold, we argue that object A and object B have \emph{hang on} relations. For example, one is classified as wall and the other is an object that can hang on the wall, like board, lamp, curtain, etc. The relations of these kinds of object pairs can be regulated as \emph{hang on}.
Such compact spatial relations are helpful for detection and understanding of the scene.

\subsection{Application of 3DRM}
\label{Strategies for backbones}
In order to apply our 3DRM to existing popular detection pipelines and verify its effectiveness and generalization, we design specific strategies to plug our 3DRM into two mainstream methods: proposal-based method and voting-based method. The detection pipeline including our Relation Module is shown in Figure \ref{fig:framework}. It is composed of two main parts, backbones for processing raw point cloud, and Relation Module for reasoning pair-wise object relations. Both the integrated and relation features are concatenated together to perform the classification and regression towards numerous candidate bounding boxes. Followed by 3D non-maximum suppression (NMS), the pipeline outputs classified and qualified 3D bounding box. We introduce how we apply 3DRM to different 3D detection frameworks as following.

\textbf{Proposal-based methods.} Lots of detection methods establish a baseline system by introducing region proposals as object candidates and classifying the objects as well as regressing the bounding box which we called proposal-based methods. These methods have achieved promising results but ignore the relation information between object candidates, so we aim to equip these methods with our 3DRM to improve the performance. Since there is no widely used framework for proposal-based methods, we design the whole backbone by ourselves. We choose over-segmentation method described in \cite{shi2019hierarchy} for raw proposal generation, object hypothesis generation module for filtering low quality proposals, PointCNN~\cite{li2018pointcnn} as feature extractor for point clouds and contextual features for enriching features of objects. Above all is the baseline for proposal-based methods in this paper.

With 3D point clouds as input, we first perform an over-segmentation on the input point cloud as described in \cite{shi2019hierarchy}.
The over-segmented patches are then merged recursively by a bottom-up fashion. The output is a binary hierarchy in which each node is a potential object (segment). We take the node segments as the initial proposals. To improve the quality of raw proposals, we first start from an object hypothesis generation module by filtering proposals with low objectness. This is achieved by using a deep neural network based on PointCNN and MLPs, predicting the objectness labels of the proposals. 

With selected and reliable proposals, each proposal can be recognized as a candidate object. Moreover, for improving the baseline performance of detection, like many of the previous works on object detection~\cite{Engelmann_2017_ICCV,Porway2008A}, we add the context information 
by exploring the points around the object candidate with radius $R$ to extract enriched contextual feature. 

Both of the original geometry features and enriched features, which extract corresponding information from the object itself and its surroundings, consist of the integrated feature for each object candidate.
After that, we apply our 3DRM to predict the pair-wise object relations and extract relation features.

Then we perform concatenation of multiple features learning from different aspects including geometric features, context features and relation features. After the concatenation, we feed the concatenated features into MLPs to predict the categorical label and regress the bounding of detection box. Finally, a 3D non-maximum suppression (NMS) is used to remove the redundant proposal candidates and obtain the final 3D objects with their bounding boxes.

\textbf{Voting-based methods.}
VoteNet~\cite{qi2019deep} is an end-to-end 3D object detection network based on a synergy of deep point set networks and Hough voting. Similar to classical Hough voting, VoteNet generates votes that lie close to objects centers, and then these votes are grouped and aggregated as clusters to generate box proposals. Each cluster can be regarded as an object candicate which can also be utilized to capture the relation information by our 3DRM. 

Specifically, with $N\times3$ point cloud as input, VoteNet first subsamples $M\times(3+C)$ seed points by a backbone network. Note that $C$ is the extended feature dimensions and $M$ is the sample number. Each seed point then goes through a voting module, predicts the offset to its object center and thus becomes a vote point for potential clusters. Furthermore, all the votes will be grouped into $K$ clusters each with dimension $(3+C)$.

At this stage, clusters with their features are sent to our 3DRM to extract the enriched $C_r$-dimensinal relation feature vector. Similar to proposal-based methods, we take the concated features $K\times(3+C+C_r)$ as the input for the following detection modules. In this way, the inference of the final 3D bounding boxes and the object classes will consider the compatibility with the relations, which makes the final prediction more reliable. In the following steps, We keep the same proposal and classify module as VoteNet to generate final 3D bounding boxes. More details will be described in the Section\ref{implementation}.

\subsection{Loss function}
\label{Loss Function}
The loss for our 3DRM is simply formulated as $\mathcal{L}_{{rn}}$ using the binary cross entropy. In this way, it is judged independently whether an object pair should have a certain relation. As for the detection pipeline, different backbones lead to diverse designs for the final loss function.

\textbf{Proposal-based methods.}
The network can be trained in an end-to-end manner with a multi-task loss including a semantic classification loss of the object candicate, a 3D bounding box regression loss and a classification loss of relations. We weigh the losses with the parameter $\lambda_1, \lambda_2, \lambda_3$ to make sure they are in similar scales. In our experiments, we set $\lambda_1=1.0, \lambda_2=10, \lambda_3=0.5$.

\begin{equation}
  \mathcal{L}_{{total}} = \lambda_1\mathcal{L}_{{cls}} + \lambda_2\mathcal{L}_{{reg}} + \lambda_3\mathcal{L}_{{rn}} 
  \label{equ:loss}
  \end{equation}

\textbf{Voting-based methods.} The network is trained in an end-to-end manner with a multi-task loss including a voting loss, an objectness loss, a 3D bounding box estimation loss, a semantic classification loss and a relation loss. It is worthy to note that the objectness loss is designed to help the proposal module to generate good enough proposals. The component losses except the voting loss are weighted by $\lambda_1, \lambda_2, \lambda_3, \lambda_4$. In our experiments, we set $\lambda_1=0.5, \lambda_2=1.0, \lambda_3=0.1, \lambda_4=0.1,$.
\begin{equation}
  \mathcal{L}_{{total}} = \mathcal{L}_{vote} + \lambda_1\mathcal{L}_{{objectness}} + \lambda_2\mathcal{L}_{{box}} + \lambda_3\mathcal{L}_{{cls}} + \lambda_4\mathcal{L}_{{rn}} 
  \label{equ:loss}
  \end{equation}

\label{implementation}
\section{Implementation Details}

In this section, we describe some implementation details of the network architectures of our 3DRM, the relevant parameters in previous methods and strategies for the application of our 3DRM to different backbones in training and testing.

\textbf{Details in 3DRM.} As mentioned in Section \ref{Relation reasoning}, there are some differences in Relation Module between \cite{santoro2017simple} and our architecture. We process a pair of objects' features at a time instead of the whole features of all objects. There are thousands of objects and the number of object pairs' permutation is too large to train. In order to obtain a stable relation feature and a faster convergence speed, we sample fixed number $k = 8$ object pairs for each object in the same scene by two ways.
One is random sampling, and the other is nearest sampling. Different sampling of object pairs results in slightly different performance (see Section~\ref{Evaluation on PointCNN baseline} and Section~\ref{Evaluation on VoteNet baseline}). For relation label computation, we set the axis-wise threshold of distance $\tau_x=\tau_y=\tau_z=0.1$ and $\tau_{xy}=\tau_{xz}=\tau_{yz}=0.5$ for  IoU threshold of bounding boxes projection onto planes. The function $ g_\theta $ and $f_\phi$ in Relation Module are different for two backbones and detailed in the following paragraphs. 

\textbf{Details in proposal-based detection.} We pre-train the object hypothesis generation module to get the object candidates. Then, we train the object relation module and detection module for final classification and regression end-to-end.
For this backbone, we use 4 fully connected layers for MLP $ g_\theta $, and 2 fully connected layers for MLP $f_\phi$ in Relation Module to extract relation features and four fully convolutional (FC) layers to predict the four types of relations. The context points around object candidates are obtained by KDTree~\cite{Bentley:1975:MBS:361002.361007} with $R=0.5m$. Finally, the geometric features, context feature and relation feature of the object are leveraged to perform the final prediction. Furthermore, there are also some differences of using data in training and test. We leverage train data whose $ IoU \geq 0.5 $ against ground truth. At test time, we use all object candidates filtered by the object hypothesis generation module. 

Note that it is hard to train object hypotheses generation module with the unbalanced training data. We use Cross Entropy Loss and two training strategies in our method: one is data balancing and the other is hard negative mining. In data balancing, we randomly choose negative samples with the same number of positive samples to form the training data. In hard negative mining, we keep the same procedure as original paper.

We implement our approach using TensorFlow. 
The Adam optimizer is leveraged in our experiments with a base learning rate of 0.001. We train the model with the maximum training epoch number as 50 and batchsize as 8 on one NVIDIA TITIAN V GPU.

\textbf{Details in voting-based detection.} The Relation Module for VoteNet backbone is slightly different from the module for proposal-based backbone. The inputs for our Relation Module are features of clusters with 128-dim. The $ g_\theta $ layer is realized through a multi-layer perceptron with FC and the output channel size is 256. The features are further processed by the MLP $f_\phi$ to get the 128-dim relation features after the channel-wise mean operation. At the same time, the outputs of $ g_\theta $ are sent to classifiers to predict the relation labels. Note that we combine four types of relations into semantic and spatial relations in voting-based detection. Each classifier predicts one type of relation and is implemented with two FCs, output of which is 128 and 2 respectively. As a result, the combined features of input features and relation features with dimension $128+128$ will be sent to the following modules.

We train the entire network end-to-end and use the same optimizer, batch size, initial learning and learning rate decay steps as VoteNet. It takes around 180 epochs for the model to converge on one NVIDIA TITAN V GPU while training.

\label{results}
\section{Experiments}

In this section, we evaluate the proposed 3DRM applied on proposal-based methods and voting-based methods respectively, in the field of 3D object detections with point cloud of indoor scenes as input. Experiments are performed on three large 3D indoor scene datasets and evaluated on the detection benchmarks. (Section~\ref{Experimental dataset}). The evaluation metric 
is described in Section~\ref{Evaluation metric}. We analyze the improved performance after applying our Relation Module on the two mentioned detection pipelines (OSegNet in Section~\ref{Evaluation on PointCNN baseline} and VoteNet in Section~\ref{Evaluation on VoteNet baseline}). Note that since we plan to verify the effectiveness and generalization of 3DRM on detection pipelines with low or relatively high performance, we choose to evaluate our 3DRM on OSegNet and VoteNet respectively. Experiments settings including evaluation on detection and ablation studies are the same for these two pipelines. Further discussion is illustrated in Section~\ref{Further discussion}.  Both of the quantitative and qualitative results demonstrate the effectiveness and generalization of the proposed Relation Module.

\subsection{Dataset and benchmarks}
\label{Experimental dataset}
We leverage a widely used dataset that provides 3D point clouds of indoor scenes: Stanford large-scale 3D Indoor Spaces Dataset S3DIS \cite{Armeni20163D} for proposal-based pipeline. S3DIS is from real scans of indoor environments which contains 3D scans from Matterport scanners in 6 areas including 271 rooms. The objects in this dataset are divided into 13 categories. We perform a k-fold cross validation across areas \cite{tchapmi2017segcloud}.

Both of ScanNetV2~\cite{dai2017} and SUN RGB-D~\cite{song2015} are leveraged to evaluate the voting-based pipeline. ScanNetV2 is an RGB-D video indoor scene dataset with richly annotated 3D reconstructed meshes. It contains about 1.5K scans annotated with both semantic segmentation and object instance labels for 18 categories. Since it doesn't provide reconstructed point clouds and oriented bounding boxes, we sample the reconstructed meshes and predict axis-aligned bounding boxes in the same way as VoteNet. 

SUN RGB-D is a large single-view RGB-D dataset for scene understanding. It contains about 10K RGB-D images captured by four different sensors with accurately annotated oriented bounding boxes for 37 object categories. Note that since it doesn't provide point cloud data, we first convert the depth images to point clouds using known camera parameters.

\subsection{Evaluation metric}
\label{Evaluation metric}
We use average precision as our evaluation metric of the detected object bounding boxes against the ground truth bounding boxes. We use two $ IoU $ thresholds as 0.5 and 0.25 respectively in our experiments. The mean average precision (mAP) is the macro-average on average precision across all test categories.

\subsection{Evaluation on proposal-based framkwork}
\label{Evaluation on PointCNN baseline}
In this section, we denote the detection framework proposed in Section~\ref{Strategies for backbones} as baseline named OSegNet which utilizes proposals generated by over-segmentation and PointCNN~\cite{li2018pointcnn} as backbones. Applying our 3DRM to OSegNet is denoted as \textbf{OSegNet+RM}. We first compare our method with OSegNet on 3D object detection. We also compare our method to state-of-the-art methods and analyze the difference and gap. After that, we conduct the ablation studies to evaluate the impact of each component in our approach. Last, we demonstrate the qualitative results of our method.

\textbf{Comparison to baseline and State-of-the-art methods.} We evaluate our method against several prior works and our baseline OSegNet which detects 3D object in indoor scenes with point cloud as input:

\begin{itemize}
\item \textbf{Sliding PointCNN~\cite{li2018pointcnn}:} A baseline which contains a PointCNN backbone and detect objects in a 3D sliding window fashion.

\item \textbf{PointNet~\cite{Charles2017PointNet}:} A method that first predicts the category of all points and then uses a breadth-first search to group nearby points with the same category.

\item \textbf{SGPN~\cite{wang2018sgpn}:} A semantic instance segmentation approach for point clouds by using an embedding learning network for point pairs. 

\item \textbf{VDRAE~\cite{shi2019hierarchy}:} A variational auto-encoder that detects 3D objects in indoor scene by using a hierarchical structure.

\end{itemize}

Table \ref{tab:table1} reports the average precision on the S3DIS dataset using 6-fold cross validation across six areas with mAP@0.5. Compared to the baseline OSegNet, our method \textbf{OSegNet+RM} obtains $\textbf{54\%}$, $\textbf{36\%}$, $\textbf{26\%}$ and $\textbf{29\%}$ increase on chair, table, sofa and mAP respectively, which proves the efficiency of our Relation Module. Note that, limited to the performance of over-segmentation method for proposal generation, 
there are very few proposals with good quality for board objects, resulting in low performance on this category. While all methods are learning-based methods and there is a lack of valid proposals on board category, our method still achieves the best performance. Especially, our method get a huge improvement (\textbf{33\%} increase) compared to the state-of-the-art on chair category and \textbf{4\%} increase on mAP, thanks to our 3DRM proposed in Section \ref{Relation reasoning} and illustrated in Figure~\ref{fig:teaser}. Moreover, without relation prediction and using only the relation features, OSegNet+RM- still surpasses OSegNet by a large margin, proving the effectiveness of the relation features extracted by our method.

\begin{table}[!t]
    \caption{Comparison of our approach against prior works and the framework OSegNet on 3D object detection. We denote OSegNet+RM as OSegNet equipped with our 3DRM. Values report average precision at mAP@0.5 on S3DIS dataset evaluated with 6-fold cross-validation on Area1$\sim$Area6.}
    \setlength{\tabcolsep}{2.1mm}{
    \centering
    \begin{tabular}{l|cccc|c}
    \toprule
    \textbf{}	& chair	& board  & table  & sofa  & mAP\\  \hline
    Sliding PointCNN & 0.36  & 0.07  & 0.39  & 0.23 & 0.26 \\
    PointNet         & 0.34  & 0.12  & 0.47  & 0.05 & 0.25 \\
    SGPN             & 0.41  & 0.13  & 0.50  & 0.07 & 0.28 \\
    VDREA            & 0.41  & \textbf{0.14}  & \textbf{0.53} & 0.43 & 0.39 \\  \hline
    OSegNet    & 0.20  & 0.01  & 0.11 & 0.25 & 0.14 \\
    OSegNet+RM           & \textbf{0.74}  & 0.01  & 0.47  & \textbf{0.51} & \textbf{0.43} \\  
    \bottomrule
    \end{tabular}
    \label{tab:table1}}
\end{table}

\begin{table}[!t]
    \centering
    \caption{Comparison of different relations of our 3DRM applied on OSegNet framework on S3DIS dataset. Experiments are trained on Area2$\sim$Area6 and tested on Area1. We denote OSegNet+RM as OSegNet equipped with our 3DRM and OSegNet+RM- as OSegNet+RM without relation prediction. Note that board category is eliminated from comparison due to poor quality of proposals.}
    \setlength{\tabcolsep}{2.4mm}{
    \begin{tabular}{l|ccc|c}
    \toprule
    \textbf{}	& chair  & table  & sofa  & mAP\\  \hline
    OSegNet    & 0.21  & 0.07 & 0.28 & 0.19 \\ \hline
    OSegNet+RM(\emph{group})      & \textbf{0.72}   & 0.36 & 0.60 & 0.56 \\
    OSegNet+RM(\emph{same as})    & 0.69   & 0.34 & 0.71 & 0.58 \\
    OSegNet+RM(\emph{support})    & 0.69   & 0.34 & 0.72 & 0.58 \\
    OSegNet+RM(\emph{hang on})    & 0.70   & 0.29  & 0.69 & 0.56 \\  \hline
    OSegNet+RM(\emph{all})    & 0.69  & \textbf{0.39} & \textbf{0.77} & \textbf{0.62} \\ \hline
    OSegNet+RM-   & 0.70  & 0.32 & 0.67 & 0.56 \\ \hline
    \end{tabular}
    \label{tab:table2}}
\end{table}

\begin{table}[!t]
    \centering
    \caption{Comparison of different selection modes of object pairs on S3DIS dataset. Experiments are trained on Area2$\sim$Area6 and tested on Area1.}
    \setlength{\tabcolsep}{2.2mm}{
    \begin{tabular}{l|ccc|c}
    \toprule
    \textbf{}	& chair	 & table  & sofa  & mAP\\  \hline
    OSegNet+RM(random)    & \textbf{0.69}   & \textbf{0.39} & \textbf{0.77} & \textbf{0.62} \\ \hline
    OSegNet+RM(nearest)   & 0.69   & 0.37 & 0.66 & 0.57 \\
    \bottomrule
    \end{tabular}
    \label{tab:table3}}
\end{table}

\begin{figure*}[htb]
    \centering
    \includegraphics[width=0.9\textwidth]{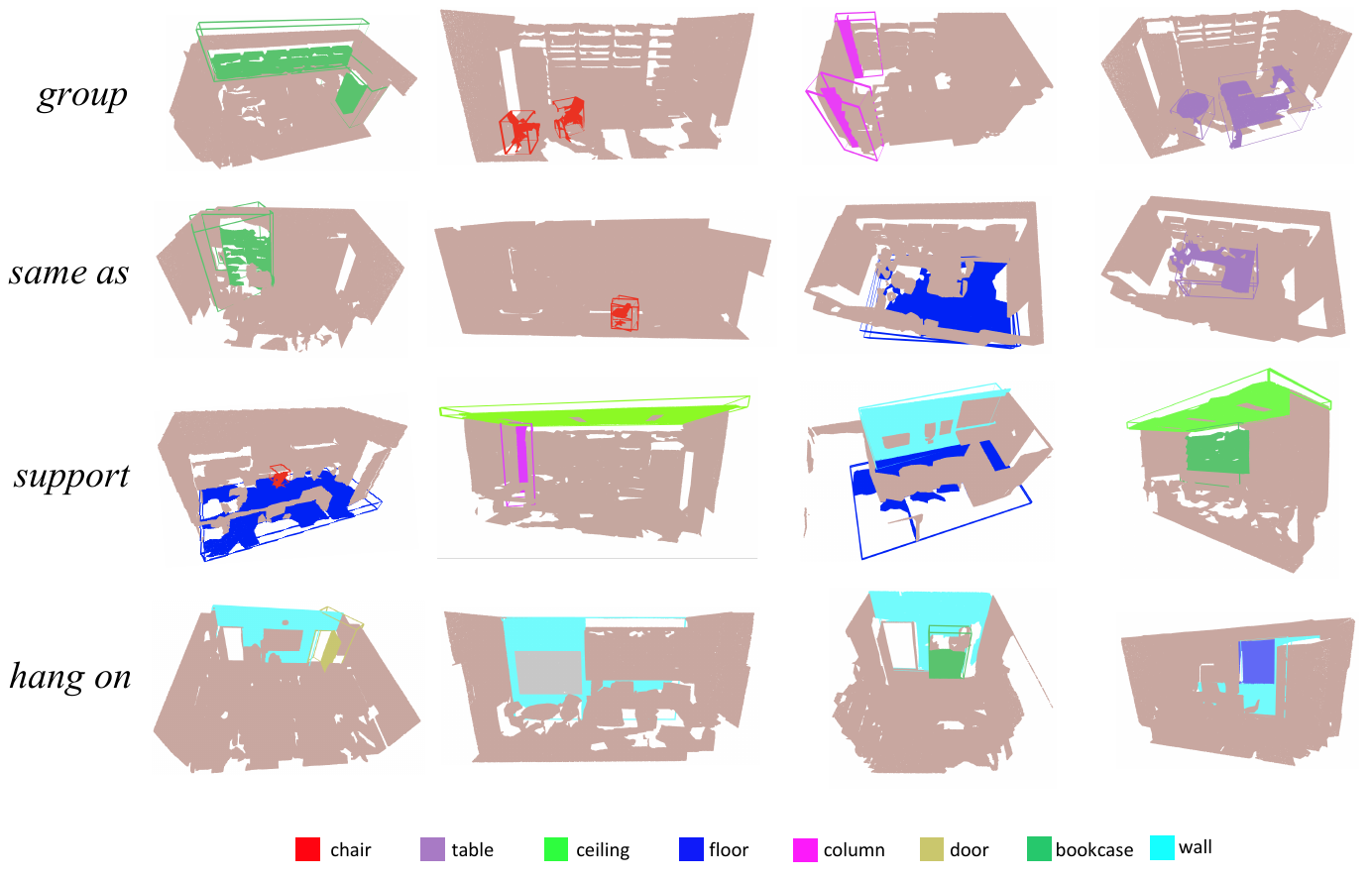}
    \caption{Visualization of the objects with semantic and spatial relationships. The first row shows the two objects with \emph{group} relation. The rest rows are for \emph{same as}, \emph{support} and \emph{hang on} relations respectively.}
    \label{fig:vis-relation2}
\end{figure*}

\textbf{Ablation study.} We evaluate the impact of each component of our approach to investigate the efficiency of different relations and relation features. Note that, all experiments for the ablation studies are trained on Area2$\sim$Area6 and tested on Area1. Board category is eliminated from comparison due to poor quality of proposals generated by OSegNet.

Specifically, we first analyze the contribution of different relations to our 3DRM. The ablation results are shown in Table~\ref{tab:table2}. While our method \textbf{OSegNet+RM}(\emph{all}) which predicts all relations at one time achieves the best performance on table and sofa with mAP of \textbf{0.62} on three categories. OSegNet+RM using only \emph{group} relations achieves the highest AP \textbf{0.72} on category of chair. We argue that, on S3DIS dataset, spatial structures of objects like table and sofa are various and complex to distinguish, leading to the dependence of both semantic and spatial relations. Shapes of chairs are relatively fixed, which relies more on semantic relations like \emph{group}.

As for the selection mode of object pairs, we also conduct the ablation study about random mode and nearest mode. Experiments are trained on Area2$\sim$Area6 and tested on Area1. Table~\ref{tab:table3} illustrates that random selection outperforms selecting object pairs according to the nearest euclidean distance on all terms. Specifically, random selection surpasses the nearest selection by \textbf{5\%} on mAP. The improvement shows that object pairs randomly selected provide more information for the network to learn, while the pairs selected nearestly can be regarded as a kind of local information.

\begin{figure*}[htb]
    \centering
    \includegraphics[width=0.98\textwidth]{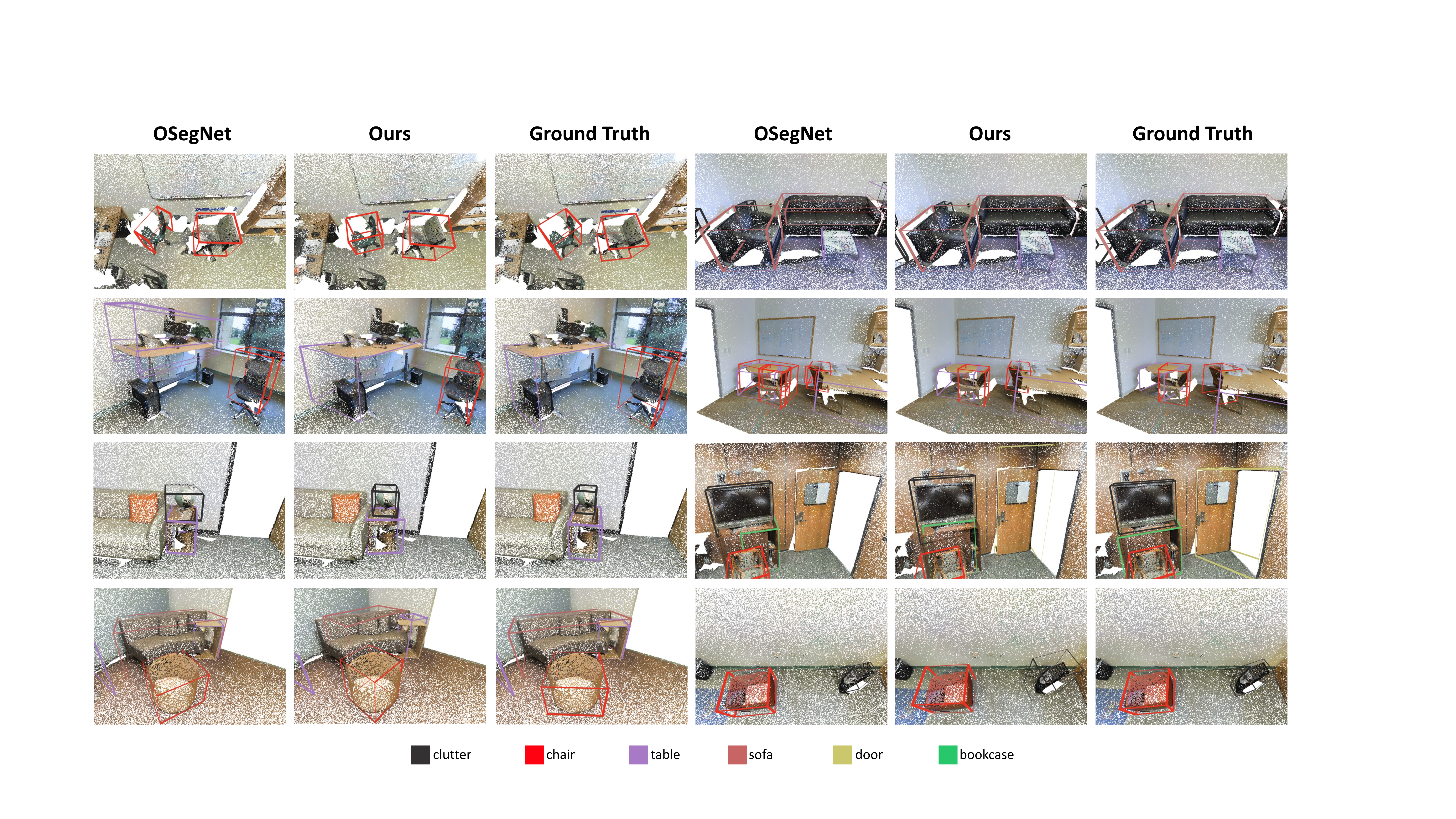}
    \caption{3D detection using our 3DRM with OSegNet on the S3DIS test set. The first/fifth column shows the bounding boxes for the OSegNet. The second/fourth column shows the qualitative detections with our 3DRM called OSegNet+RM  and the third/sixth column shows the ground truth of that. Our method is capable of detecting objects in cluttered scenes and regressed bounding boxes are more accurate and distinct than OSegNet.}
    \label{fig:vis-detection-succ}
\end{figure*}

\begin{table}[!t]
    \centering
    \caption{Comparison of our approach against VoteNet on 3D object detection on ScanNetV2 val set and SUN RGB-D val set. We denote VoteNet+RM as our approach with applying our 3DRM on VoteNet.}
    \scalebox{0.95}{
    \setlength{\tabcolsep}{1.0mm}{
    \begin{tabular}{l|cc|cc}
    \toprule
    \textbf{}   &\multicolumn{2}{c|}{mAP@0.25}      &\multicolumn{2}{c}{mAP@0.5} \\
    \textbf{}		& ScanNet & SUN RGB-D  & ScanNet & SUN RGB-D\\ \hline 
    VoteNet      & 58.6  & 57.7  & 33.5  & 33.7 \\  
    VoteNet+RM & \textbf{59.7}  & \textbf{59.1} & \textbf{37.3} & \textbf{35.1}\\  
    \bottomrule
    \end{tabular}
    \label{tab:votenet-all}}}
\end{table}



\textbf{Visualization of relation prediction.} Figure \ref{fig:vis-relation2} visualizes the relations between objects in the same room. Both of semantic and spatial relations provide rich context information to help detection. Different objects belonging to the same categories implies that they should have similar shapes. If objects belonging to the same instance have overlapping bounding boxes, this helps the network to regress the box correctly. For spatial relations, such explicit information provides extra knowledge to accomplish the task of detection.

\textbf{Qualitative examples.} Figure \ref{fig:vis-detection-succ} shows several qualitative results of object detection on S3DIS dataset. The proposed Relation Module leverages multiple features from relations to help detection module classify the objects in 3D bounding boxes. We visualize the result of OSegNet (first column and fourth column), our method OSegNet+RM (second column and fifth column) and ground truth (third column and last column). It is demonstrated that our method is capable of detecting objects in cluttered scenes and regressed bounding boxes are more accurate and distinct than OSegNet.

\subsection{Evaluation on voting-based framework}
\label{Evaluation on VoteNet baseline}
We first compare our method with the baseline VoteNet on 3D object detection in 3D point clouds. Results justify the effectiveness and practicality of the proposed 3DRM. After that, we conduct extensive ablation studies to evaluate the impact of each component in our approach. Lastly, we demonstrate the qualitative results of our method.

\begin{table*}[!t]
    \centering
    \caption{Comparison to VoteNet with mAP@0.5 on ScanNetV2 val set for our method with different relations. We denote VoteNet+RM as VoteNet equipped with our 3DRM and VoteNet+RM- as VoteNet+RM without relation prediction.}
    \scalebox{0.84}{
    \setlength{\tabcolsep}{0.82mm}{
    \begin{tabular}{l|cccccccccccccccccc|c}
    \toprule
    \textbf{}	& wind	& bed  & cntr & sofa & tabl & showr & ofurn & sink & pic & chair & desk & curt & fridge & door & toil & bkshf & bath & cab & mAP\\  \hline
    VoteNet         & 7.89  & 76.70  & 20.11  & 69.04  & 41.80  & 7.75  & 14.05  & 21.06  & 0.76  & 67.30  & 32.52  & 10.58  & 28.89  & 14.68  & 82.07  & 27.86  & 79.31  & 9.36  & 33.99 \\ \hline
    VoteNet+RM(semantic) & \textbf{12.29}  & 80.63  & 14.59  & \textbf{71.79}  & 41.28  & 10.41  & 13.35  & \textbf{29.46}  & 0.14  & 67.73  & 34.74  & 16.95  & \textbf{37.79}  & 15.70  & \textbf{89.96}  & \textbf{44.22}  & 82.95  & 8.03  & \textbf{37.33} \\ 
    VoteNet+RM(spatial) & 10.48  & 81.17  & \textbf{22.07}  & 68.95  & 42.02  & 4.20  & \textbf{16.56}  & 26.08  & \textbf{1.57}  & 68.83  & 36.49  & 13.30  & 33.02  & \textbf{17.67}  & 84.37  & 39.66 & \textbf{89.43}  & \textbf{10.79}  & 37.04 \\ 
    VoteNet+RM(all) & 9.68  & 77.97  & 21.72  & 67.65  & 41.89  & \textbf{15.11}  & 14.56  & 26.58  & 0.22  & \textbf{69.20}  & \textbf{40.30}  & \textbf{26.30}  & 30.37  & 13.67  & 89.89  & 32.41 & 78.94  & 9.63  & 37.01 \\ \hline
    VoteNet+RM- & 10.33  & \textbf{81.40}  & 18.97  & 66.57  & \textbf{42.94}  & 9.33  & 15.45  & 25.55  & 0.42  & 69.13  & 37.72  & 18.83  & 29.75  & 15.20  & 87.40  & 40.66 & 82.26  & 6.61  & 36.58 \\ 
    \bottomrule
    \end{tabular}}}
    \label{tab:votenet-relations-05}
\end{table*}

\begin{table}[!t]
    \centering
    \caption{Comparison of different selection modes of object pairs on ScanNetV2 val set. We denote VoteNet+RM as our approach with applying our 3DRM on VoteNet.}
    \setlength{\tabcolsep}{2.5mm}{
    \begin{tabular}{l|cc}
    \toprule
    \textbf{}	  	& mAP@0.25  & mAP@0.5\\  \hline
    VoteNet         & 58.6  & 33.5 \\ \hline
    VoteNet+RM(random)  & \textbf{59.73}   & \textbf{37.33}  \\
    VoteNet+RM(nearest) & 58.44  & 36.79 \\  
    \bottomrule
    \end{tabular}
    \label{tab:votenet-ranornear}}
\end{table}

\textbf{Comparison to VoteNet.} We evaluate our method against VoteNet in 3D object detection. Quantitative results on ScanNet and SUN RGB-D are summarized in Table~\ref{tab:votenet-all}. We apply the proposed Relation Module to the representative VoteNet and denote the network as \textbf{VoteNet+RM} as our method. Note that we take the performance VoteNet+RM with semantic relations only as the final results since it performs the best on these two datasets. Our method significantly outperforms VoteNet by not only \textbf{1.1\%} and \textbf{3.8\%} on ScanNet, but also \textbf{1.4\%} and \textbf{1.4\%} on SUN RGB-D in terms of mAP with IoU=0.25 and IoU=0.5 respectively. Note that, our method increase the performance of VoteNet by $3.8\%$ on mAP@0.5 which illustrates that our 3DRM can not only mitigate ambiguity but also increase accuracy of the detection. Furthermore, we argue that this benefits from the enriched relation features from our 3DRM, which provides comprehensive understanding to the object and its surrounding environment. More quantitative results are shown in appendix.

\begin{figure*}[htb]
    \centering
    \includegraphics[width=0.8\textwidth]{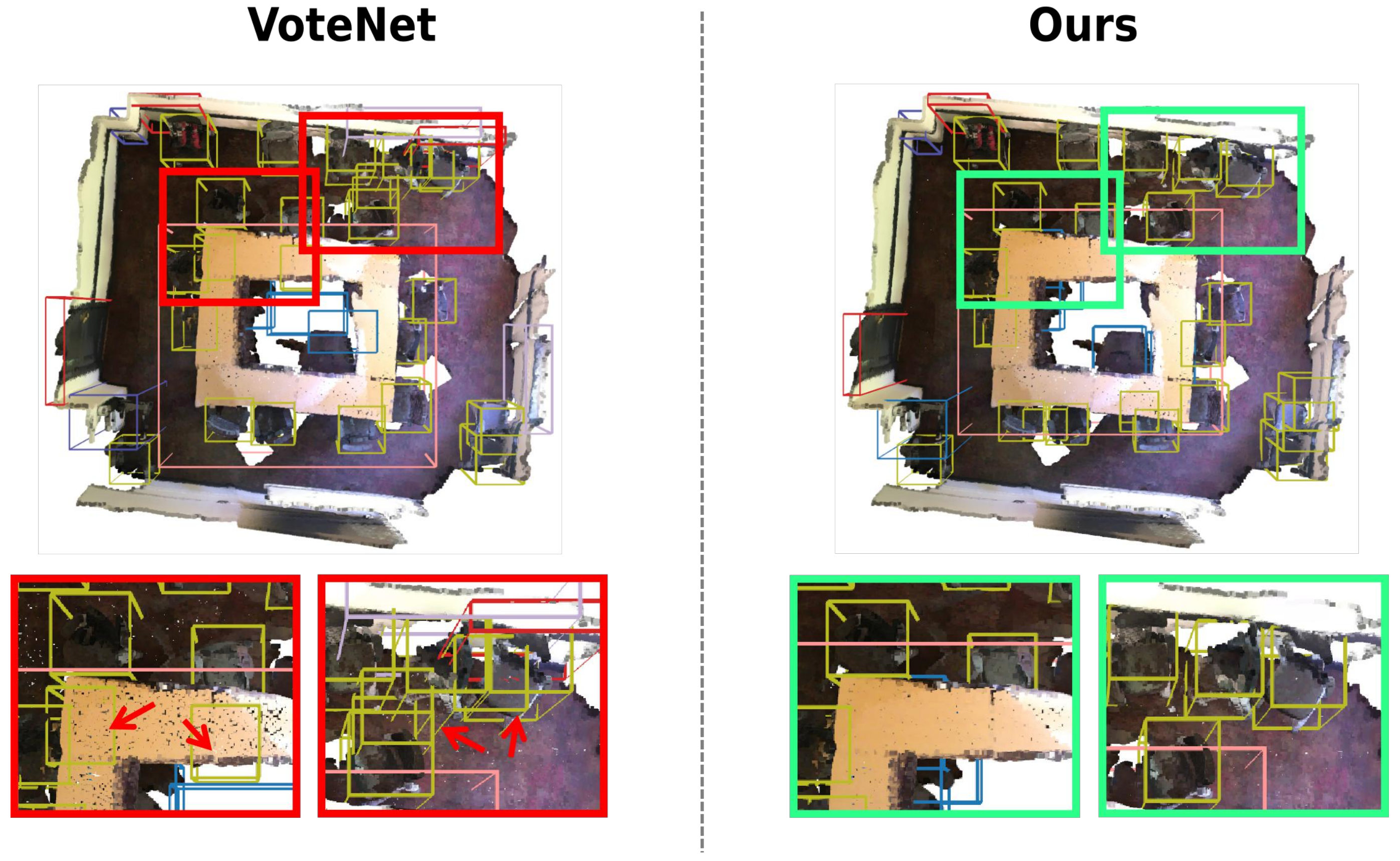}
    \caption{Qualitative comparison results of 3D object detection on ScanNetV2 val set. Left: VoteNet, Right: Ours. The detailed comparison demonstrates that our 3DRM enables more accurate and reasonable detection. Color is for depiction, not used for detection. }
    \label{fig:vis-scannet}
\end{figure*}

\textbf{Ablation study.} Extensive ablation studies are performed to verify the increased accuracy of our approach. Note that we combine \emph{group} and \emph{same as} relations as semantic relations and \emph{support} and \emph{hang on} as spatial relations. To prove the efficiency of the proposed 3DRM, we first study how the semantic and spatial relations help the task of 3D object detection for different categories. Applying our 3DRM to VoteNet without predicting relation labels denoted as VoteNet+RM- is also considered. Results on ScanNet are shown in Table~\ref{tab:votenet-relations-05}.

The results show that VoteNet+RM with only semantic relations achieves the best performance with an increase of \textbf{3.34\%} in terms of mAP@0.5. The reason is that  objects of most categories are sensitive to semantic relations, and can obtain more context information from semantic relations. For example, objects, such as windows, sofas, fridges, toilets and so on, have simple and clear spatial structure, and thus need various objects and context to help understanding. Objects, such as counters, pictures, doors, baths and cabs, are usually placed in a complex environment where semantic information is rich enough and spatial information are critical to them since their structures are more complicated. Moreover, some categories of objects benefit from both semantic and spatial relations like shower, chair, desk, etc. Note that applying our 3DRM to VoteNet without predicting relation labels and only use the relation features also outperforms VoteNet, which proves that relation features extracted by our 3DRM do help detect 3D objects better.

As for the mode of selecting object pairs, we compare two ways of selection: random mode and nearest mode. Comparison results are illustrated in Table~\ref{tab:votenet-ranornear}, It is clear that random selection of object pairs achieves higher performance than selecting several nearest objects to form relation pairs. This is because random selection can provide various object pairs distributed in the whole scene and thus enrich the information around objects to improve the detection quality.

\textbf{Qualitative results and discussion.} The qualitative results on ScanNet are shown in Figure~\ref{fig:vis-scannet}. Ours method detects the objects more accurately and robustly, which is beneficial from our 3DRM. It is noteworthy that detection results of VoteNet are confused with other objects and ambiguous in some areas with noisy point cloud, while ours can classify and locate the objects precisely and clearly without redundant bounding boxes. We argue that this is attributed to the pair-wise relation reasoning. Details are shown in the second row of Figure~\ref{fig:vis-scannet} where red rectangles on the left refer to the ambiguous and wrong detections on chairs by VoteNet. Green rectangles on the right demonstrate the accurate detections by ours.

\begin{figure*}[!t]
  \centering
  \includegraphics[width=1.0\textwidth]{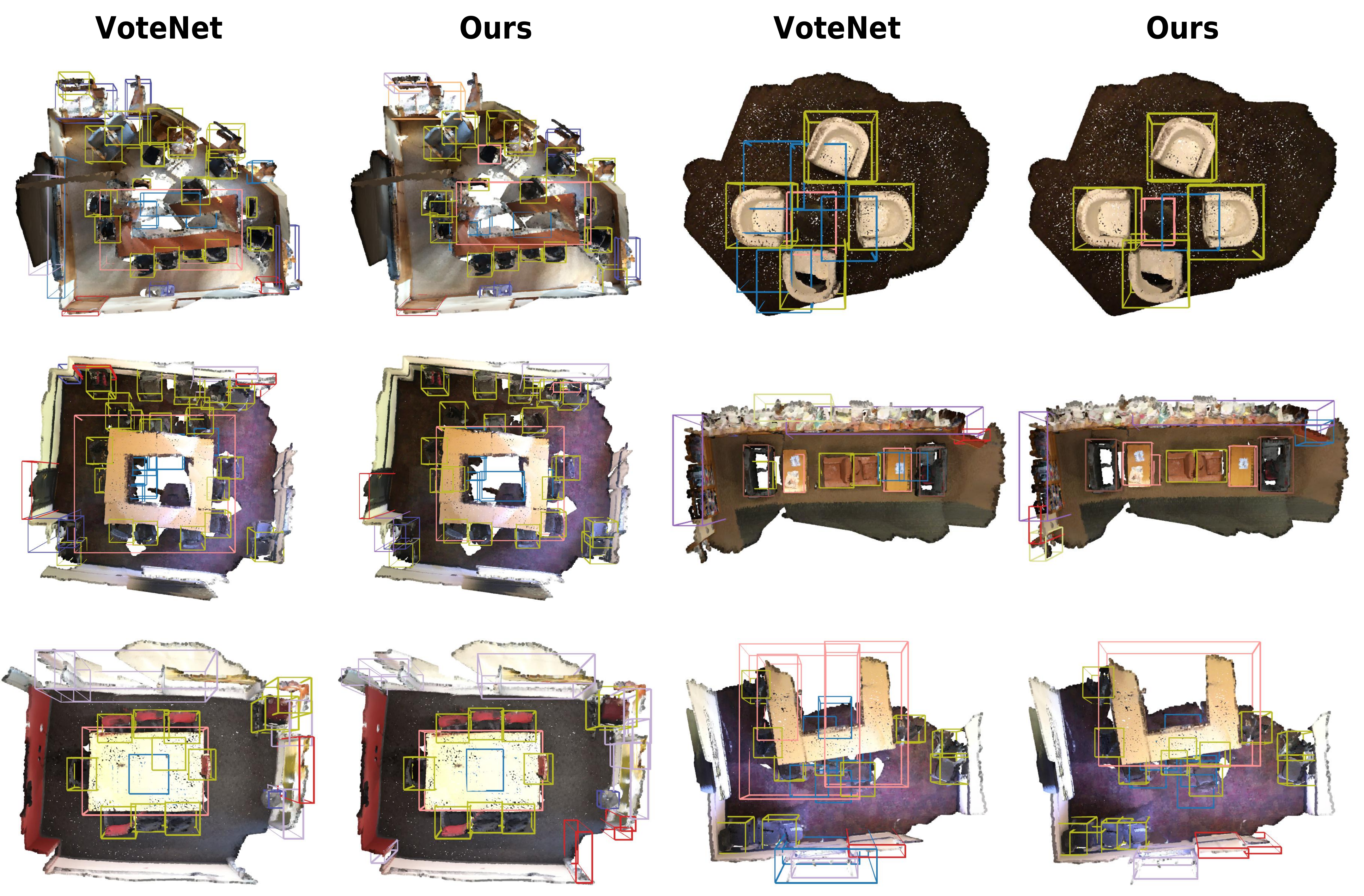}
  \caption{3D detection using our 3DRM with VoteNet on the ScanNetV2 val set. The first/third column shows the bounding boxes for the VoteNet. The second/fourth column shows the qualitative detections with our 3DRM called VoteNet+RM.}
  \label{fig:vis-detection-scannet}
\end{figure*}

Figure~\ref{fig:vis-detection-scannet} shows the detection results on ScanNetV2 val set. From the comparison of Ours and VoteNet, we can detect the objects accurately and robustly with less ambiguity. Specifically, in some cluttered areas, ours can distinguish different objects and regress the bounding boxes precisely. There are usually many chairs in scenes like offices and it is quite common to misunderstand the chairs as other categories due to noise and their various appearance. Our 3DRM can help alleviate this problem by using relation reasoning and thus achieve better detection results.

\begin{figure}[!t]
    \centering
    \includegraphics[width=0.37\textwidth]{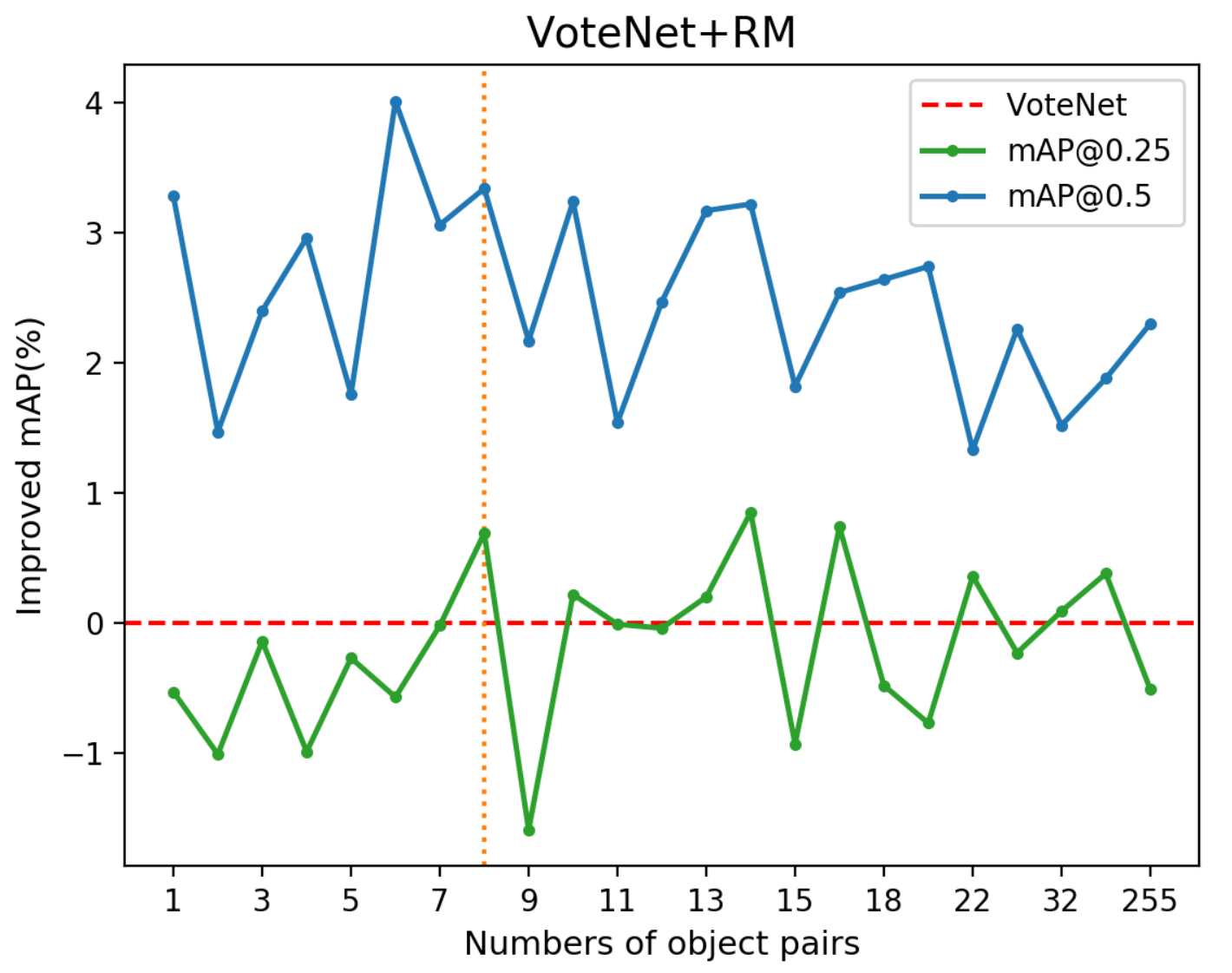}
    \caption{Improved percentage of mAP for different numbers of object pairs. We denote VoteNet+RM as VoteNet equipped with our 3DRM. Sampling 8 object pairs for an object achieves the most improvement taking both mAP@0.25 and mAP@0.5 as well as computational efficiency into consideration.}
    \label{fig:num_of_pairs}
\end{figure}

\subsection{Further discussion}
\label{Further discussion}
\textbf{Complexity and computational efficiency.} Our 3DRM is consisted of semantic relations and spatial relations including \emph{group}, \emph{same as}, \emph{support} and \emph{hang on}. Among these, relations like \emph{group}, \emph{same as} are easy and fast to compute since we only need to compare semantic or instance labels. The computational time for semantic relations can be ignored. As for spatial relations, Algorithm~\ref{algorithm1} can illustrate the complexity. Note that although we use loops for simplication in the algorithm, we actually use matrix multiplication for further acceleration in experiments. Moreover, the computational time and complexity grows as the number of object pairs is larger. Numerically, we test the computational efficiency of spatial relations for 2048 proposals in ScanNet dataset. We sample 8 object pairs for each proposal, which means $8*2048$ object pairs in total. The total time is around $0.047s$ and the average time is approximately $3\times10^{-6} s$ for each object pair, which proves the high efficiency of our 3DRM.

With regard to the number of object pairs for each object which may have influence on the complexity of relation computation, we perform experiments by comparing the contributions of different pair numbers to the detection with VoteNet+RM on ScanNet dataset. Results are shown in Figure~\ref{fig:num_of_pairs}. Explicitly, sampling 8 or 14 object pairs for an object achieves the most improvement considering both mAP@0.25 and mAP@0.5. Since these contributions of these two numbers of pairs are almost the same, we build 8 object pairs for each one for the balance of computational efficiency and performance.

\textbf{Improvement on different pipelines.} We have evaluated the improved performance of detection on mAP for both OSegNet and VoteNet to justify the generalization of our 3DRM on detection pipelines with different basic performance. Explicitly, applying 3DRM to OSegNet obtains \textbf{29\%} improvement on mAP@0.5 on S3DIS while VoteNet gets \textbf{3.8\%} on ScanNet and \textbf{1.4\%} on SUN RGB-D. OSegNet is an intuitively simple baseline implemented mainly with over-segmentation for proposal generation and PointCNN as backbones. Initial proposals generated by over-segmentation in OSegNet are relatively less organized and accurate compared to VoteNet which relies on deep hough voting. The network architecture of OSegNet is much simpler than VoteNet. Less organized proposals and simpler architecture result in lower basic performance for OSegNet. However, this actually demonstrates the generalization and effectiveness of our 3DRM by being able to help attain a huge improvement on detection for simple pipelines, and comparable improvement even for comprehensive pipelines.

\label{conclusion}
\section{Conclusions}

We presented a Relation Module for 3D object detection on large-scale scene datasets. With the object candidates generated from backbones, we predict object relations and capture relation features by our 3DRM, which is capable of mitigating the ambiguity of 3D object detection, thus helping locate and classify the 3D objects more accurately and robustly. We applied our 3DRM to both proposal-based methods and voting-based methods. Improved detection results demonstrate the effectiveness and generalization of our method.

\textbf{Limitations.} Although experiments verify the effectiveness of our 3DRM, our method can only predict pair-wise object relations explicitly. High-level relations that may help scene understanding are not considered.

\textbf{Future work.} There are several directions worth trying for the future work. First, it is worth trying to add more relation types to the relation module. Second, we only perform relation reasoning on object pairs. To explore the possibility to apply relation networks for analyzing sub-scenes is an interesting direction. Finally, to perform relation reasoning in more complicated 3D task (such as Vision Question Answering on 3D scenes) is also a promising direction.

{\small
\bibliographystyle{ieee_fullname}
\bibliography{egbib}

\begin{thebibliography}{10}\itemsep=-1pt

\bibitem{Agrawal2017VQA}
Aishwarya Agrawal, Jiasen Lu, Stanislaw Antol, Margaret Mitchell, C.~Lawrence
  Zitnick, Devi Parikh, and Dhruv Batra.
\newblock Vqa: Visual question answering.
\newblock {\em International Journal of Computer Vision}, 123(1):4--31, 2017.

\bibitem{Armeni20163D}
Iro Armeni, Ozan Sener, Amir~R Zamir, Helen Jiang, Ioannis Brilakis, Martin
  Fischer, and Silvio Savarese.
\newblock 3d semantic parsing of large-scale indoor spaces.
\newblock In {\em Proceedings of the IEEE Conference on Computer Vision and
  Pattern Recognition}, pages 1534--1543, 2016.

\bibitem{Bentley:1975:MBS:361002.361007}
Jon~Louis Bentley.
\newblock Multidimensional binary search trees used for associative searching.
\newblock {\em Commun. ACM}, 18(9):509--517, Sept. 1975.

\bibitem{cadene2019murel}
Remi Cadene, Hedi Ben-Younes, Matthieu Cord, and Nicolas Thome.
\newblock Murel: Multimodal relational reasoning for visual question answering.
\newblock In {\em Proceedings of the IEEE Conference on Computer Vision and
  Pattern Recognition}, pages 1989--1998, 2019.

\bibitem{Chen_2020_CVPR}
Jintai Chen, Biwen Lei, Qingyu Song, Haochao Ying, Danny~Z. Chen, and Jian Wu.
\newblock A hierarchical graph network for 3d object detection on point clouds.
\newblock In {\em Proceedings of the IEEE/CVF Conference on Computer Vision and
  Pattern Recognition (CVPR)}, pages 392--401, June 2020.

\bibitem{chen2017spatial}
Xinlei Chen and Abhinav Gupta.
\newblock Spatial memory for context reasoning in object detection.
\newblock In {\em Proceedings of the IEEE International Conference on Computer
  Vision}, pages 4086--4096, 2017.

\bibitem{Chen2017Multi}
Xiaozhi Chen, Huimin Ma, Ji Wan, Bo Li, and Tian Xia.
\newblock Multi-view 3d object detection network for autonomous driving.
\newblock In {\em Proceedings of the IEEE Conference on Computer Vision and
  Pattern Recognition}, pages 1907--1915, 2017.

\bibitem{cui2020learning}
Qiongjie Cui, Huaijiang Sun, and Fei Yang.
\newblock Learning dynamic relationships for 3d human motion prediction.
\newblock In {\em Proceedings of the IEEE/CVF Conference on Computer Vision and
  Pattern Recognition}, pages 6519--6527, 2020.

\bibitem{dai2017}
Angela Dai, Angel~X Chang, Manolis Savva, Maciej Halber, Thomas Funkhouser, and
  Matthias Nie{\ss}ner.
\newblock Scannet: Richly-annotated 3d reconstructions of indoor scenes.
\newblock In {\em Proceedings of the IEEE Conference on Computer Vision and
  Pattern Recognition}, pages 5828--5839, 2017.

\bibitem{duan2019structural}
Yueqi Duan, Yu Zheng, Jiwen Lu, Jie Zhou, and Qi Tian.
\newblock Structural relational reasoning of point clouds.
\newblock In {\em Proceedings of the IEEE Conference on Computer Vision and
  Pattern Recognition}, pages 949--958, 2019.

\bibitem{Engelmann20CVPR}
Francis Engelmann, Martin Bokeloh, Alireza Fathi, Bastian Leibe, and Matthias
  Nie{\ss}ner.
\newblock 3d-mpa: Multi-proposal aggregation for 3d semantic instance
  segmentation.
\newblock In {\em Proceedings of the IEEE/CVF Conference on Computer Vision and
  Pattern Recognition}, pages 9031--9040, 2020.

\bibitem{Engelmann_2017_ICCV}
Francis Engelmann, Theodora Kontogianni, Alexander Hermans, and Bastian Leibe.
\newblock Exploring spatial context for 3d semantic segmentation of point
  clouds.
\newblock In {\em Proceedings of the IEEE International Conference on Computer
  Vision Workshops}, pages 716--724, 2017.

\bibitem{fan2020fsod}
Qi Fan, Wei Zhuo, Chi-Keung Tang, and Yu-Wing Tai.
\newblock Few-shot object detection with attention-rpn and multi-relation
  detector.
\newblock In {\em Proceedings of the IEEE/CVF Conference on Computer Vision and
  Pattern Recognition}, pages 4013--4022, 2020.

\bibitem{feng2020deep}
Di Feng, Christian Haase-Schuetz, Lars Rosenbaum, Heinz Hertlein, Claudius
  Glaeser, Fabian Timm, Werner Wiesbeck, and Klaus Dietmayer.
\newblock Deep multi-modal object detection and semantic segmentation for
  autonomous driving: Datasets, methods, and challenges.
\newblock {\em IEEE Transactions on Intelligent Transportation Systems}, 2020.

\bibitem{feng2020relation}
Mingtao Feng, Syed~Zulqarnain Gilani, Yaonan Wang, Liang Zhang, and Ajmal Mian.
\newblock Relation graph network for 3d object detection in point clouds.
\newblock {\em IEEE Transactions on Image Processing}, 30:92--107, 2020.

\bibitem{Hou_2019_CVPR}
Ji Hou, Angela Dai, and Matthias Niessner.
\newblock 3d-sis: 3d semantic instance segmentation of rgb-d scans.
\newblock In {\em Proceedings of the IEEE/CVF Conference on Computer Vision and
  Pattern Recognition (CVPR)}, pages 4421--4430, June 2019.

\bibitem{hu2018relation}
Han Hu, Jiayuan Gu, Zheng Zhang, Jifeng Dai, and Yichen Wei.
\newblock Relation networks for object detection.
\newblock In {\em Proceedings of the IEEE Conference on Computer Vision and
  Pattern Recognition}, pages 3588--3597, 2018.

\bibitem{huang2018cooperative}
Siyuan Huang, Siyuan Qi, Yinxue Xiao, Yixin Zhu, Ying~Nian Wu, and Song-Chun
  Zhu.
\newblock Cooperative holistic scene understanding: Unifying 3d object, layout,
  and camera pose estimation.
\newblock In {\em Advances in Neural Information Processing Systems}, pages
  207--218, 2018.

\bibitem{Huang_2018_ECCV}
Siyuan Huang, Siyuan Qi, Yixin Zhu, Yinxue Xiao, Yuanlu Xu, and Song-Chun Zhu.
\newblock Holistic 3d scene parsing and reconstruction from a single rgb image.
\newblock In {\em Proceedings of the European Conference on Computer Vision
  (ECCV)}, pages 187--203, September 2018.

\bibitem{huang2016structure}
Shi-Sheng Huang, Hongbo Fu, and Shi-Min Hu.
\newblock Structure guided interior scene synthesis via graph matching.
\newblock {\em Graphical Models}, 85:46--55, 2016.

\bibitem{huang2015support}
Shi-Sheng Huang, Hongbo Fu, Ling-Yu Wei, and Shi-Min Hu.
\newblock Support substructures: Support-induced part-level structural
  representation.
\newblock {\em IEEE transactions on visualization and computer graphics},
  22(8):2024--2036, 2015.

\bibitem{huang2020improving}
Yifei Huang, Yusuke Sugano, and Yoichi Sato.
\newblock Improving action segmentation via graph-based temporal reasoning.
\newblock In {\em Proceedings of the IEEE/CVF Conference on Computer Vision and
  Pattern Recognition}, pages 14024--14034, 2020.

\bibitem{Krishna2016Visual}
Ranjay Krishna, Yuke Zhu, Oliver Groth, Justin Johnson, Kenji Hata, Joshua
  Kravitz, Stephanie Chen, Yannis Kalantidis, Li~Jia Li, and David~A. Shamma.
\newblock Visual genome: Connecting language and vision using crowdsourced
  dense image annotations.
\newblock {\em International Journal of Computer Vision}, 123(1):32--73, 2016.

\bibitem{ku2018joint}
Jason Ku, Melissa Mozifian, Jungwook Lee, Ali Harakeh, and Steven~L Waslander.
\newblock Joint 3d proposal generation and object detection from view
  aggregation.
\newblock In {\em 2018 IEEE/RSJ International Conference on Intelligent Robots
  and Systems (IROS)}, pages 1--8. IEEE, 2018.

\bibitem{kulkarni20193d}
Nilesh Kulkarni, Ishan Misra, Shubham Tulsiani, and Abhinav Gupta.
\newblock 3d-relnet: Joint object and relational network for 3d prediction.
\newblock In {\em Proceedings of the IEEE International Conference on Computer
  Vision}, pages 2212--2221, 2019.

\bibitem{le2020hierarchical}
Thao~Minh Le, Vuong Le, Svetha Venkatesh, and Truyen Tran.
\newblock Hierarchical conditional relation networks for video question
  answering.
\newblock In {\em Proceedings of the IEEE/CVF Conference on Computer Vision and
  Pattern Recognition}, pages 9972--9981, 2020.

\bibitem{li2020spatial}
Xia Li, Yibo Yang, Qijie Zhao, Tiancheng Shen, Zhouchen Lin, and Hong Liu.
\newblock Spatial pyramid based graph reasoning for semantic segmentation.
\newblock In {\em Proceedings of the IEEE/CVF Conference on Computer Vision and
  Pattern Recognition}, pages 8950--8959, 2020.

\bibitem{li2018pointcnn}
Yangyan Li, Rui Bu, Mingchao Sun, Wei Wu, Xinhan Di, and Baoquan Chen.
\newblock Pointcnn: Convolution on x-transformed points.
\newblock In {\em Advances in neural information processing systems}, pages
  820--830, 2018.

\bibitem{Liang2018Deep}
Ming Liang, Bin Yang, Shenlong Wang, and Raquel Urtasun.
\newblock {\em Deep Continuous Fusion for Multi-sensor 3D Object Detection}.
\newblock 2018.

\bibitem{Lin_2013_ICCV}
Dahua Lin, Sanja Fidler, and Raquel Urtasun.
\newblock Holistic scene understanding for 3d object detection with rgbd
  cameras.
\newblock In {\em Proceedings of the IEEE International Conference on Computer
  Vision (ICCV)}, pages 1417--1424, December 2013.

\bibitem{liu2020beyond}
Chenchen Liu, Yang Jin, Kehan Xu, Guoqiang Gong, and Yadong Mu.
\newblock Beyond short-term snippet: Video relation detection with
  spatio-temporal global context.
\newblock In {\em Proceedings of the IEEE/CVF Conference on Computer Vision and
  Pattern Recognition}, pages 10840--10849, 2020.

\bibitem{liu2019relation}
Yongcheng Liu, Bin Fan, Shiming Xiang, and Chunhong Pan.
\newblock Relation-shape convolutional neural network for point cloud analysis.
\newblock In {\em Proceedings of the IEEE Conference on Computer Vision and
  Pattern Recognition}, pages 8895--8904, 2019.

\bibitem{mi2020hierarchical}
Li Mi and Zhenzhong Chen.
\newblock Hierarchical graph attention network for visual relationship
  detection.
\newblock In {\em Proceedings of the IEEE/CVF Conference on Computer Vision and
  Pattern Recognition}, pages 13886--13895, 2020.

\bibitem{mou2019relation}
Lichao Mou, Yuansheng Hua, and Xiao~Xiang Zhu.
\newblock A relation-augmented fully convolutional network for semantic
  segmentation in aerial scenes.
\newblock In {\em Proceedings of the IEEE conference on computer vision and
  pattern recognition}, pages 12416--12425, 2019.

\bibitem{Porway2008A}
Jake Porway, Kristy Wang, Benjamin Yao, and Song~Chun Zhu.
\newblock A hierarchical and contextual model for aerial image understanding.
\newblock In {\em Computer Vision and Pattern Recognition, 2008. CVPR 2008.
  IEEE Conference on}, 2008.

\bibitem{Qi_2019_ICCV}
Charles~R. Qi, Or Litany, Kaiming He, and Leonidas~J. Guibas.
\newblock Deep hough voting for 3d object detection in point clouds.
\newblock In {\em Proceedings of the IEEE/CVF International Conference on
  Computer Vision (ICCV)}, pages 9277--9286, October 2019.

\bibitem{qi2019deep}
Charles~R Qi, Or Litany, Kaiming He, and Leonidas~J Guibas.
\newblock Deep hough voting for 3d object detection in point clouds.
\newblock In {\em Proceedings of the IEEE International Conference on Computer
  Vision}, pages 9277--9286, 2019.

\bibitem{Qi_2018_CVPR}
Charles~R Qi, Wei Liu, Chenxia Wu, Hao Su, and Leonidas~J Guibas.
\newblock Frustum pointnets for 3d object detection from rgb-d data.
\newblock In {\em Proceedings of the IEEE conference on computer vision and
  pattern recognition}, pages 918--927, 2018.

\bibitem{Charles2017PointNet}
Charles~R Qi, Hao Su, Kaichun Mo, and Leonidas~J Guibas.
\newblock Pointnet: Deep learning on point sets for 3d classification and
  segmentation.
\newblock In {\em Proceedings of the IEEE conference on computer vision and
  pattern recognition}, pages 652--660, 2017.

\bibitem{qi2017pointnet++}
Charles~Ruizhongtai Qi, Li Yi, Hao Su, and Leonidas~J Guibas.
\newblock Pointnet++: Deep hierarchical feature learning on point sets in a
  metric space.
\newblock {\em Advances in neural information processing systems},
  30:5099--5108, 2017.

\bibitem{qi20173d}
Xiaojuan Qi, Renjie Liao, Jiaya Jia, Sanja Fidler, and Raquel Urtasun.
\newblock 3d graph neural networks for rgbd semantic segmentation.
\newblock In {\em Proceedings of the IEEE International Conference on Computer
  Vision}, pages 5199--5208, 2017.

\bibitem{santoro2017simple}
Adam Santoro, David Raposo, David~G Barrett, Mateusz Malinowski, Razvan
  Pascanu, Peter Battaglia, and Timothy Lillicrap.
\newblock A simple neural network module for relational reasoning.
\newblock In {\em Advances in neural information processing systems}, pages
  4967--4976, 2017.

\bibitem{shi2019pointrcnn}
Shaoshuai Shi, Xiaogang Wang, and Hongsheng Li.
\newblock Pointrcnn: 3d object proposal generation and detection from point
  cloud.
\newblock In {\em Proceedings of the IEEE Conference on Computer Vision and
  Pattern Recognition}, pages 770--779, 2019.

\bibitem{shi2019hierarchy}
Yifei Shi, Angel~X Chang, Zhelun Wu, Manolis Savva, and Kai Xu.
\newblock Hierarchy denoising recursive autoencoders for 3d scene layout
  prediction.
\newblock In {\em Proceedings of the IEEE Conference on Computer Vision and
  Pattern Recognition}, pages 1771--1780, 2019.

\bibitem{shi2016data}
Yifei Shi, Pinxin Long, Kai Xu, Hui Huang, and Yueshan Xiong.
\newblock Data-driven contextual modeling for 3d scene understanding.
\newblock {\em Computers \& Graphics}, 55:55--67, 2016.

\bibitem{Simonyan2014Two}
Karen Simonyan and Andrew Zisserman.
\newblock Two-stream convolutional networks for action recognition in videos.
\newblock {\em Advances in neural information processing systems}, 27:568--576,
  2014.

\bibitem{song2017web3d}
Peihua Song, Youyi Zheng, and Jinyuan Jia.
\newblock Web3d learning platform of furniture layout based on case-based
  reasoning and distance field.
\newblock In {\em International Conference on Technologies for E-Learning and
  Digital Entertainment}, pages 235--250. Springer, 2017.

\bibitem{song2015}
Shuran Song, Samuel~P Lichtenberg, and Jianxiong Xiao.
\newblock Sun rgb-d: A rgb-d scene understanding benchmark suite.
\newblock In {\em Proceedings of the IEEE conference on computer vision and
  pattern recognition}, pages 567--576, 2015.

\bibitem{sung2018learning}
Flood Sung, Yongxin Yang, Li Zhang, Tao Xiang, Philip~HS Torr, and Timothy~M
  Hospedales.
\newblock Learning to compare: Relation network for few-shot learning.
\newblock In {\em Proceedings of the IEEE Conference on Computer Vision and
  Pattern Recognition}, pages 1199--1208, 2018.

\bibitem{tchapmi2017segcloud}
Lyne Tchapmi, Christopher Choy, Iro Armeni, JunYoung Gwak, and Silvio Savarese.
\newblock Segcloud: Semantic segmentation of 3d point clouds.
\newblock In {\em 2017 international conference on 3D vision (3DV)}, pages
  537--547. IEEE, 2017.

\bibitem{wang2019exploring}
Wenbin Wang, Ruiping Wang, Shiguang Shan, and Xilin Chen.
\newblock Exploring context and visual pattern of relationship for scene graph
  generation.
\newblock In {\em Proceedings of the IEEE Conference on Computer Vision and
  Pattern Recognition}, pages 8188--8197, 2019.

\bibitem{wang2018sgpn}
Weiyue Wang, Ronald Yu, Qiangui Huang, and Ulrich Neumann.
\newblock Sgpn: Similarity group proposal network for 3d point cloud instance
  segmentation.
\newblock In {\em Proceedings of the IEEE Conference on Computer Vision and
  Pattern Recognition}, pages 2569--2578, 2018.

\bibitem{Xie_2020_CVPR}
Qian Xie, Yu-Kun Lai, Jing Wu, Zhoutao Wang, Yiming Zhang, Kai Xu, and Jun
  Wang.
\newblock Mlcvnet: Multi-level context votenet for 3d object detection.
\newblock In {\em Proceedings of the IEEE/CVF Conference on Computer Vision and
  Pattern Recognition (CVPR)}, pages 10447--10456, June 2020.

\bibitem{Xu_2019_CVPR}
Hang Xu, Chenhan Jiang, Xiaodan Liang, and Zhenguo Li.
\newblock Spatial-aware graph relation network for large-scale object
  detection.
\newblock In {\em Proceedings of the IEEE Conference on Computer Vision and
  Pattern Recognition}, pages 9298--9307, 2019.

\bibitem{Yang2018PIXOR}
Bin Yang, Wenjie Luo, and Raquel Urtasun.
\newblock Pixor: Real-time 3d object detection from point clouds.
\newblock In {\em Proceedings of the IEEE conference on Computer Vision and
  Pattern Recognition}, pages 7652--7660, 2018.

\bibitem{yang2019learning}
Bo Yang, Jianan Wang, Ronald Clark, Qingyong Hu, Sen Wang, Andrew Markham, and
  Niki Trigoni.
\newblock Learning object bounding boxes for 3d instance segmentation on point
  clouds.
\newblock In {\em Advances in Neural Information Processing Systems}, pages
  6737--6746, 2019.

\bibitem{Ye_2018_ECCV}
Xiaoqing Ye, Jiamao Li, Hexiao Huang, Liang Du, and Xiaolin Zhang.
\newblock 3d recurrent neural networks with context fusion for point cloud
  semantic segmentation.
\newblock In {\em Proceedings of the European Conference on Computer Vision
  (ECCV)}, pages 403--417, 2018.

\bibitem{zhang2017deepcontext}
Yinda Zhang, Mingru Bai, Pushmeet Kohli, Shahram Izadi, and Jianxiong Xiao.
\newblock Deepcontext: Context-encoding neural pathways for 3d holistic scene
  understanding.
\newblock In {\em Proceedings of the IEEE International Conference on Computer
  Vision}, pages 1192--1201, 2017.

\bibitem{deepcontext}
Yinda Zhang, Mingru Bai, Pushmeet Kohli, Shahram Izadi, and Jianxiong Xiao.
\newblock Deepcontext: Context-encoding neural pathways for 3d holistic scene
  understanding.
\newblock In {\em 2017 IEEE International Conference on Computer Vision
  (ICCV)}, pages 1201--1210, 10 2017.

\bibitem{zhang2020relation}
Zhizheng Zhang, Cuiling Lan, Wenjun Zeng, Xin Jin, and Zhibo Chen.
\newblock Relation-aware global attention for person re-identification.
\newblock In {\em Proceedings of the IEEE/CVF Conference on Computer Vision and
  Pattern Recognition}, pages 3186--3195, 2020.

\bibitem{H3DNet}
Zaiwei Zhang, Bo Sun, Haitao Yang, and Qixing Huang.
\newblock H3dnet: 3d object detection using hybrid geometric primitives.
\newblock In Andrea Vedaldi, Horst Bischof, Thomas Brox, and Jan-Michael Frahm,
  editors, {\em Proceedings of the European Conference on Computer Vision
  (ECCV)}, pages 311--329, Cham, 2020. Springer International Publishing.

\bibitem{zhao2018triangle}
Yawei Zhao, Kai Xu, En Zhu, Xinwang Liu, Xinzhong Zhu, and Jianping Yin.
\newblock Triangle lasso for simultaneous clustering and optimization in graph
  datasets.
\newblock {\em IEEE Transactions on Knowledge and Data Engineering},
  31(8):1610--1623, 2018.

\bibitem{zheng2014recurring}
Youyi Zheng, Daniel Cohen-Or, Melinos Averkiou, and Niloy~J Mitra.
\newblock Recurring part arrangements in shape collections.
\newblock In {\em Computer Graphics Forum}, volume~33, pages 115--124. Wiley
  Online Library, 2014.

\bibitem{zheng2020foreground}
Zhuo Zheng, Yanfei Zhong, Junjue Wang, and Ailong Ma.
\newblock Foreground-aware relation network for geospatial object segmentation
  in high spatial resolution remote sensing imagery.
\newblock In {\em Proceedings of the IEEE/CVF Conference on Computer Vision and
  Pattern Recognition}, pages 4096--4105, 2020.

\end{thebibliography}
}

\end{document}